# Deep Huber quantile regression networks

Hristos Tyralis[1,2,*], Georgia Papacharalampous[3], Nilay Dogulu[4], Kwok P. Chun[5]

[1]Department of Topography, School of Rural, Surveying and Geoinformatics Engineering, National Technical University of Athens, Iroon Polytechniou 5, 157 80 Zografou, Greece (montchrister@gmail.com, hristos@itia.ntua.gr, https://orcid.org/0000-0002-8932-4997)

[2]Construction Agency, Hellenic Air Force, Mesogion Avenue 227–231, 15 561 Cholargos, Greece (montchrister@gmail.com, hristos@itia.ntua.gr, https://orcid.org/0000-0002-8932-4997)

[3]Department of Topography, School of Rural, Surveying and Geoinformatics Engineering, National Technical University of Athens, Iroon Polytechniou 5, 157 80 Zografou, Greece (papacharalampous.georgia@gmail.com, https://orcid.org/0000-0001-5446-954X)

[4]Hydrology, Water Resources and Cryosphere Branch, World Meteorological Organisation (WMO), Geneva, Switzerland (ndogulu@wmo.int, nilay.dogulu@gmail.com, https://orcid.org/0000-0003-4229-2788)

[5]Department of Geography and Environmental Management, University of the West of England, Bristol, United Kingdom (kwok.chun@uwe.ac.uk, https://orcid.org/0000-0001-9873-6240)

*Corresponding author

**Abstract**: Typical machine learning regression applications aim to report the mean or the median of the predictive probability distribution, via training with a squared or an absolute error scoring function. The importance of issuing predictions of more functionals of the predictive probability distribution (quantiles and expectiles) has been recognized as a means to quantify the uncertainty of the prediction. In deep learning (DL) applications, that is possible through quantile and expectile regression neural networks (QRNN and ERNN respectively). Here we introduce deep Huber quantile regression networks (DHQRN) that nest QRNN and ERNN as edge cases. DHQRN can predict Huber quantiles, which are more general functionals in the sense that they nest quantiles and expectiles as limiting cases. The main idea is to train a DL algorithm with the Huber quantile scoring function, which is consistent for the Huber quantile functional. As a proof of concept, DHQRN are applied to predict house prices in Melbourne, Australia and Boston, United States (US). In this context, predictive performances of three DL architectures are discussed along with evidential interpretation of results from two economic case studies. Additional simulation experiments and applications to real-world case studies using open datasets demonstrate a satisfactory absolute performance of DHQRN.

## 1. Introduction

Predictions of models should be probabilistic, taking the form of probability distributions (Dawid 1984, Gneiting and Raftery 2007) to quantify their uncertainty. Although the vast majority of deep learning (DL) regression algorithms is designed for issuing point predictions, interest in probabilistic predictions continues to increase, with related methods having been reviewed by Khosravi et al. (2011), Kabir et al. (2018), Abdar et al. (2021), Hüllermeier and Waegeman (2021), Zhou et al. (2022), Tyralis and Papacharalampous (2024). Applications of probabilistic predictions with machine learning (ML) can be found in various scientific fields including, e.g., energy research (Hong and Fan 2016), environmental sciences and remote sensing (Papacharalampous et al. 2020, 2024, Papacharalampous and Tyralis 2022), and forecasting in econometrics and finance (Tyralis and Papacharalampous 2024) among others.

Two major classes of ML probabilistic prediction algorithms exist. The first class of algorithms is of parametric type, in the sense that the probability distribution of the dependent variable is specified first. In the second stage, the algorithm estimates the parameters of the probability distribution. Algorithms in this class have mostly been influenced by techniques proposed by Rigby and Stasinopoulos (2005), and related developments (Stasinopoulos et al. 2018). They include extensions of Generalized Additive Models for Location, Scale and Shape (GAMLSS) models to general ML algorithms. The second major class includes non-parametric methods (e.g., quantile and expectile regression) reviewed by Kneib (2013), Kneib et al. (2023) and Tyralis and Papacharalampous (2024) among others. In the latter class, algorithms are designed to optimize a suitable scoring function (also called loss function, e.g., the quantile or the expectile scoring function). Subsequently, the trained algorithm can issue multiple point predictions at various quantile (or expectile) levels. A dense grid of those predictions at multiple quantile (or expectile) levels can characterize the probability distribution of the prediction. Similar ideas in non-parametric settings have also been proposed for training environmental models (e.g. Pande 2013a, 2013b, Tyralis and Papacharalampous 2021b, Tyralis et al. 2023).

A class of non-parametric ML algorithms for probabilistic prediction includes neural networks and DL in general. Respective DL algorithms include quantile regression neural networks (QRNN, Taylor 2000) and expectile regression neural networks (ERNN, Jiang et al. 2017). Variants of QRNN have also been developed by Cannon (2011), Xu et al. (2016), Xu et al. (2017), Jantre et al. (2021), Moon et al. (2021), Jia and Jeong (2022) among others.

The general class of Huber quantiles introduced by Taggart (2022b) includes the classes of quantiles and expectiles as limiting cases. Quantiles, expectiles and Huber quantiles share some important properties. In particular, they are elicitable functionals (Gneiting 2011, for the definition of elicitability see Section 2.3), while they have their economic interpretation in investment problems (Ehm et al. 2016). Huber quantiles are more general compared to quantiles and expectiles, since they offer an additional degree of flexibility that allows answering research questions associated to investment problems, where caps on profits and losses are imposed (for the importance of imposing caps, see the environmental as well as the finance applications in Taggart 2022b). On the other hand, an economic interpretation of quantiles is related to binary betting and the cost-loss decision problem (Ehm et al. 2016, Taggart 2022b).

The aim of this manuscript is to introduce a new class of DL regression that issues point predictions of Huber quantiles. We call it *deep Huber quantile regression networks* (DHQRN). QRNN and ERNN are limiting cases of DHQRN. The main concept of DHQRN is using as objective function, the generalized Huber quantile scoring function (Taggart 2022b) that is consistent for Huber quantiles (the definition of consistency of scoring functions can be found later in Section 2.3). The motivation to introduce DHQRN comes from (a) the importance of Huber quantiles as general quantiles as well as integral parts of probabilistic forecasting and prediction applications, and (b) the advantage of DL as an effective approach for extracting data representations. Huber quantiles also belong to the general class of $M$-quantiles (Breckling and Chambers 1988). Therefore, the problem of training DHQRN falls to the conditional $M$-quantile estimation category (Dimitriadis et al. 2024), while related DL conditional $M$-quantile estimation developments are limited. To understand various aspects of DHQRN, we include comparisons of three DL architectures with skill scores (Gneiting 2011) as well as the economic interpretation of the results, in an application to house price prediction. ML for house price prediction is an evolving topic (Antipov and Pokryshevskaya 2012, Dimopoulos et al. 2018, Dimopoulos and Bakas 2019) that also includes the case of quantile regression (McMillen 2008) when probabilistic predictions are required.

The remainder of the manuscript is structured as follows: Section 2 provides definitions of quantiles, expectiles and Huber quantiles, and presents the concepts of elicitability and consistency that are especially relevant for characterizing functionals and associated scoring functions. The skill scores for comparing the algorithms and a general theory for the economic interpretation is also included. The proposed DL algorithm (DHQRN) as well as QRNN and ERNN are described in Section 3, while two applications of house price predictions, a simulation experiment, as well four applications using open datasets are presented in Section 4. The manuscript ends with concluding remarks in Section 5.

## 2. Methods

Section 2.1 provides a brief overview of the methodology, with subsequent sections providing formalizations. In Section 2.2, we present the concept of statistical functionals of probability distributions (in particular, quantile, expectile and Huber quantile statistical functionals are presented). We also describe, in Section 2.3, the concept of elicitability that is a favorable property of quantiles, expectiles and Huber quantiles and the concept of consistency, which refers to favorable properties of scoring functions that are used to assess predictions of elicitable statistical functionals. Section 2.4 discusses economic interpretation of scoring functions with particular focus on house investment decisions.

### 2.1 Methodology overview

The fundamentals of key concepts that the methodology is based on are summarized here briefly since these concepts are relatively new to the DL community. In particular, the problem resolved in the article can be decomposed into two elements: 1) the concepts of elicitable functionals and consistent scoring functions, and 2) their specific applications in DL that extend the traditional applications. Before proceeding to the methodology overview, we provide some plain language descriptions for some new terms for the DL community adapted from Taggart (2022a); formal definitions and original sources are provided in the following Sections.

**Functional**: A property of a probability distribution, e.g. the mean, the median, the quantile.

**Scoring function**: A function that assigns a numerical score to a pair of predictions-observations, e.g. the squared error scoring function.

**Consistent scoring function**: If the modeler receives a directive to predict a functional, it is critical that the scoring function for the functional is consistent, i.e. the expected score is minimized when following the directive.

**Strictly consistent scoring function**: A scoring function is strictly consistent if it is consistent and its expected score is uniquely minimized at the true value of the functional.

**Elicitable functional**: A functional for which a strictly consistent scoring function exists.

To be more specific, the goal of deep Huber quantile regression networks is to predict Huber quantiles using DL. Huber quantiles are elicitable functionals, meaning that a strictly consistent scoring function should be used as a loss function in regression settings so that the algorithm is trained to predict Huber quantiles (Dimitriadis et al. 2024). In particular, the Huber quantile scoring function is strictly consistent for Huber quantiles. Elicitability and strict consistency are related concepts; in the sense, that if a functional is elicitable, then there is a strictly consistent function for it, and vice versa. By assigning specific values to the parameters of Huber quantiles and Huber quantile scoring functions, one can obtain their respective edge (limiting) cases, which include the quantile functional and the quantile scoring function, as well as the expectile functional and the expectile scoring function.

Consequently, to develop a DL regression algorithm that can predict Huber quantiles, it suffices for the optimization algorithm to use a Huber quantile scoring function as the objective function to train the DL algorithm (Dimitriadis et al. 2024). A summary of the methodology is shown in Figure 1. Figure 1 contains three boxes, each of which represents a different concept. Box (a) contains the Huber quantile, along with its two edge cases;

namely the quantile and the expectile (as well as their respective special cases, i.e. the median and the mean). Box (b) contains the Huber quantile scoring function, along with its two edge cases; namely the quantile scoring function and the expectile scoring function (as well as their respective special cases, i.e. the absolute error scoring function and the squared error scoring function). Box (c) contains deep Huber quantile regression networks and its edge cases. The arrow connecting the three boxes represents the relationship between these concepts. The Huber quantile scoring function is employed as the objective function of deep Huber quantile regression networks, which can then be used for predicting Huber quantiles. In practice, to predict Huber quantiles, one should use a Huber quantile scoring function as an objective function when training a deep learning regression model. This will ensure that the model is able to learn the relationship between the input data and the desired Huber quantiles. While Figure 1 does not explicitly show the relationships between the quantile functional, the quantile scoring function, and quantile regression neural networks, these relationships are straightforward to devise. For reasons of brevity, the relationships corresponding to the expectile, median and mean functionals are not shown either.

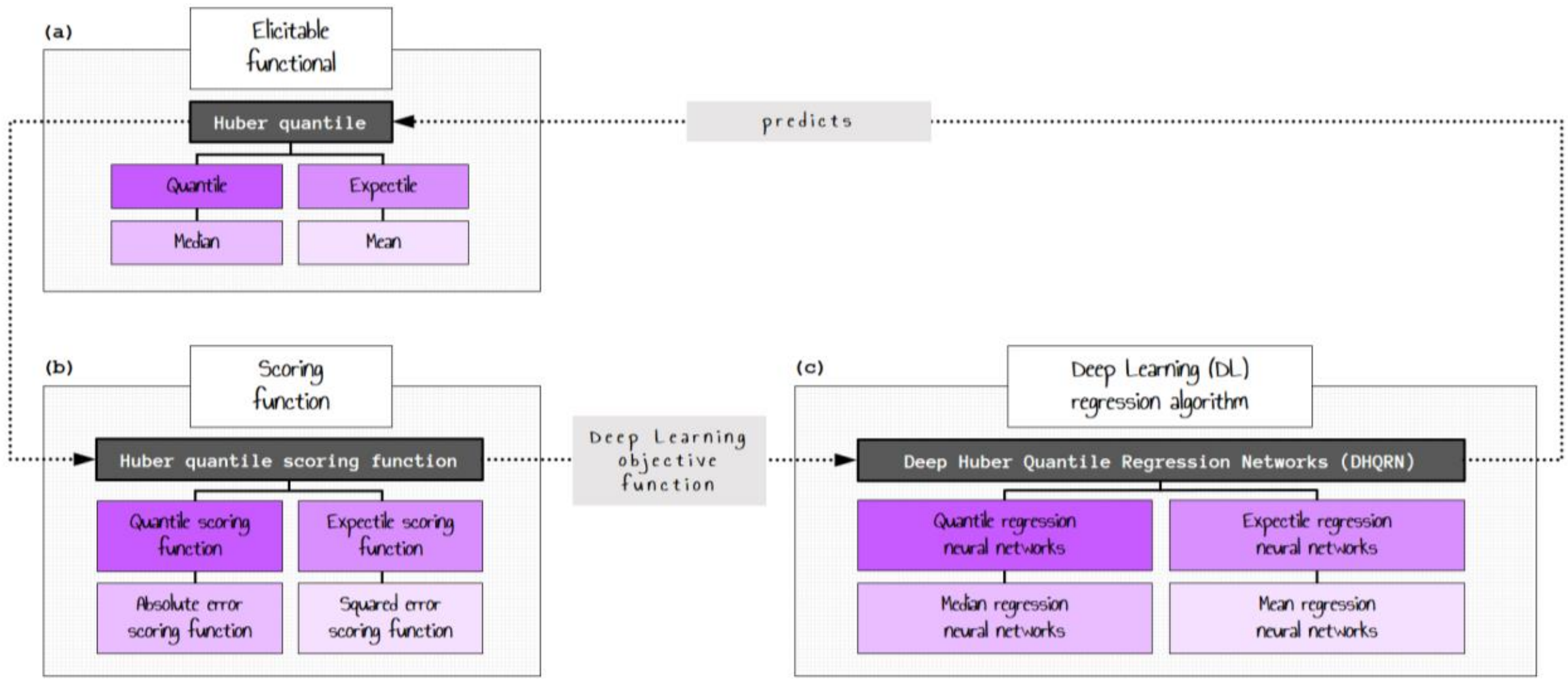

Figure 1. Overview of methodology. The boxes show: (a) the concept of Huber quantiles and its edge cases, (b) the Huber quantile scoring function and its edge cases, and (c) deep Huber quantile regression networks and its edge cases. The black dotted arrows represent the connections between concepts for the case of Huber quantiles.

## 2.2 Quantiles, expectiles and Huber quantiles

This part is intended to provide an overview of quantiles, expectiles and Huber quantiles. Please refer to Gneiting (2011) and for further information on quantile and expectile functionals (and respective consistent scoring functions) and to Taggart (2022b) for results on Huber quantiles (and respective consistent scoring functions).

*2.2.1 Definitions*

Let $\underline{y}$ be a random variable. Hereinafter, random variables will be underlined. A materialization of $\underline{y}$ will be denoted by $y$. We write $\underline{y} \sim F$ to indicate that $\underline{y}$ has cumulative distribution function (CDF) $F$, i.e.

$$F(y) := P(\underline{y} \leq y) \tag{1}$$

A statistical functional (or simply a functional) $T$ is a mapping

$$T: \mathcal{F}(I) \to \mathcal{P}(I), I \subseteq \mathbb{R} \tag{2}$$

where $\mathcal{F}(\mathbb{R})$ denotes the class of probability measures on the Borel-Lebesque sets of $\mathbb{R}$, $\mathcal{F}(I)$ denotes the subset of probability measures on $I$ and $\mathcal{P}(I)$ is the power set of $I$. An example of a statistical functional is the $\tau$-quantile $Q^{\tau}(F)$ (or quantile at level $\tau$, Koenker and Bassett Jr 1978), which is a mapping:

$$Q^{\tau}: \mathcal{F}(I) \to \mathcal{P}(I), I \subseteq \mathbb{R} \tag{3}$$

and is defined by:

$$Q^{\tau}(F) = \{x \in I: \lim_{y \uparrow x} F(y) \leq \tau \leq F(x)\}, I \subseteq \mathbb{R}, \tau \in (0,1), F \in \mathcal{F}(I) \tag{4}$$

where $\mathcal{F}(I)$ is the family of probability measures on $I$ with finite first moment. The restriction on the first moment of the probability distributions of the family $\mathcal{F}(I)$ is necessary, to have elicitable quantile functionals (for the definition of elicitability of functionals, see Section 2.3).

Another well-known functional is the $\tau$-expectile $E^{\tau}(F)$ (Newey and Powell 1987, Bellini et al. 2014, Bellini and Di Bernardino 2017), which is a mapping:

$$E^{\tau}: \mathcal{F}_1(I) \to \mathcal{P}(I) \tag{5}$$

and is defined by:

$$E^{\tau}(F) = \left\{x \in I: \tau \int_{x}^{\infty} (y - x) \mathrm{d}F(y) = (1 - \tau) \int_{-\infty}^{x} (x - y) \mathrm{d}F(y)\right\}, F \in \mathcal{F}_1(I) \tag{6}$$

or equivalently

$$E^{\tau}(F) = \{x \in I: \tau \mathbb{E}_F[\kappa_{0,\infty}(\underline{y} - x)] = (1 - \tau) \mathbb{E}_F[\kappa_{0,\infty}(x - \underline{y})]\}, F \in \mathcal{F}_1(I) \tag{7}$$

where $\mathcal{F}_1(I)$ is the family of probability measures on $I$ with finite second moment, $\mathbb{E}_F$ denotes the expectation of a random variable, and $\kappa_{a,b}(t)$ is the capping function, defined by:

$$\kappa_{a,b}(t) = \max\{\min\{t, b\}, -a\} \forall t \in \mathbb{R}, a, b \in [0, \infty] \tag{8}$$

or equivalently

$$\kappa_{a,b}(t) = \begin{cases} -a, t \leq -a \\ t, -a < t \leq b \\ b, t > b \end{cases} \tag{9}$$

The $\tau$-Huber quantile functional $H^{\tau}_{a,b}(F)$ (Taggart 2022b) is a mapping

$$H^{\tau}_{a,b}: \mathcal{F}(I) \to \mathcal{P}(I) \tag{10}$$

and is defined by:

$$H^{\tau}_{a,b}(F) = \{x \in I: \tau \mathbb{E}_F[\kappa_{0,a}(\underline{y} - x)] = (1 - \tau)\mathbb{E}_F[\kappa_{0,b}(x - \underline{y})]\} \tag{11}$$

*2.2.2 Interpretations*

When the set $Q^{\tau}(F)$ is single-valued, then from eq. (4):

$$\tau = P(\underline{y} \leq Q^{\tau}(F)) \tag{12}$$

The sample equivalent of eq. (12) implies that the frequency of observations of $\underline{y}$ below the $\tau$-quantile $Q^{\tau}(F)$ is equal to $\tau$. For instance, the median of $F$ is equal to $Q^{1/2}(F)$, while the frequency of observations of $\underline{y}$ below $Q^{1/2}(F)$ is equal to 50%. Quantiles at a dense grid of quantile levels can characterize the probability distribution of the random variable $\underline{y}$.

Eq. (7) can be re-written as:

$$\tau = \mathbb{E}_F[|\underline{y} - E^{\tau}(F)| \mathbb{1}(\underline{y} \leq E^{\tau}(F))] / \mathbb{E}_F[|\underline{y} - E^{\tau}(F)|] \tag{13}$$

where $\mathbb{1}(A)$ is the indicator function (equal to 1 when the event $A$ realizes and 0 otherwise). Comparing eqs. (12) and (13), one sees that $E^{\tau}(F)$ is such that the mean distance from all $\underline{y}$ below $E^{\tau}(F)$ is $100\tau\%$ of the mean distance between $\underline{y}$ and $E^{\tau}(F)$ (Daouia et al. 2018), while $Q^{\tau}(F)$ is the point below which $100\tau\%$ of the mass of $\underline{y}$ lies. Consequently, an intuitive interpretation of the expectile is that it is similar to the quantile when replacing frequency with distance. A special case of an expectile is $E^{1/2}(F)$, which is the mean of the probability distribution $F$.

Setting $a$ and $b$ equal to $\infty$ and by combining eqs. (7) and (11) one gets $H^{\tau}_{\infty,\infty}(F) = E^{\tau}(F)$. That means that expectiles are limiting cases of expectiles. When $a = b$ and $\tau = 1/2$, the special case $H^{1/2}_{a,a}(F)$ is called Huber mean, and is related to the Huber scoring function (Huber 1964) as explained later in Section 2.3. For a single valued $\tau$-

Huber quantile functional, eq, (11) can be re-written as:

$$\tau = \mathbb{E}_F[\kappa_{0,b}(H^{\tau}_{a,b}(F) - \underline{y})]/(\mathbb{E}_F[\kappa_{0,a}(\underline{y} - H^{\tau}_{a,b}(F))] + \mathbb{E}_F[\kappa_{0,b}(H^{\tau}_{a,b}(F) - \underline{y})]) \quad (14)$$

Eq. (14) is similar to eq. (13), with the difference being that distances inside expectations are capped by the parameters $a, b$ of the Huber quantile.

Huber quantiles belong to the larger family of $M$-quantiles (Breckling and Chambers 1988), while certain $M$-quantiles (including expectiles) are related to quantiles (Jones 1994) with explicit equations. A geometric interpretation of quantiles, expectiles and Huber quantiles can be found in Taggart (2022b). Here we adapt the explanation for the case of a log-normal probability distribution with density function:

$$f(y; \mu, \sigma) := \frac{1}{y\, \sigma \sqrt{2\pi}} \exp\left(-\frac{(\log y - \mu)^2}{2\sigma^2}\right), y > 0, \mu \in \mathbb{R}, \sigma > 0 \quad (15)$$

Figure 2 shows the log-normal cumulative distribution function based on the house price empirical discussion in Section 4.1 with specified parameters $\mu = -0.063$ and $\sigma = 0.534$ to illustrate the geometric representation of eq. (15). In the specific problem, the random variable $\underline{y}$ is Melbourne house prices measured in $10^6$ \$ (Australian dollars).

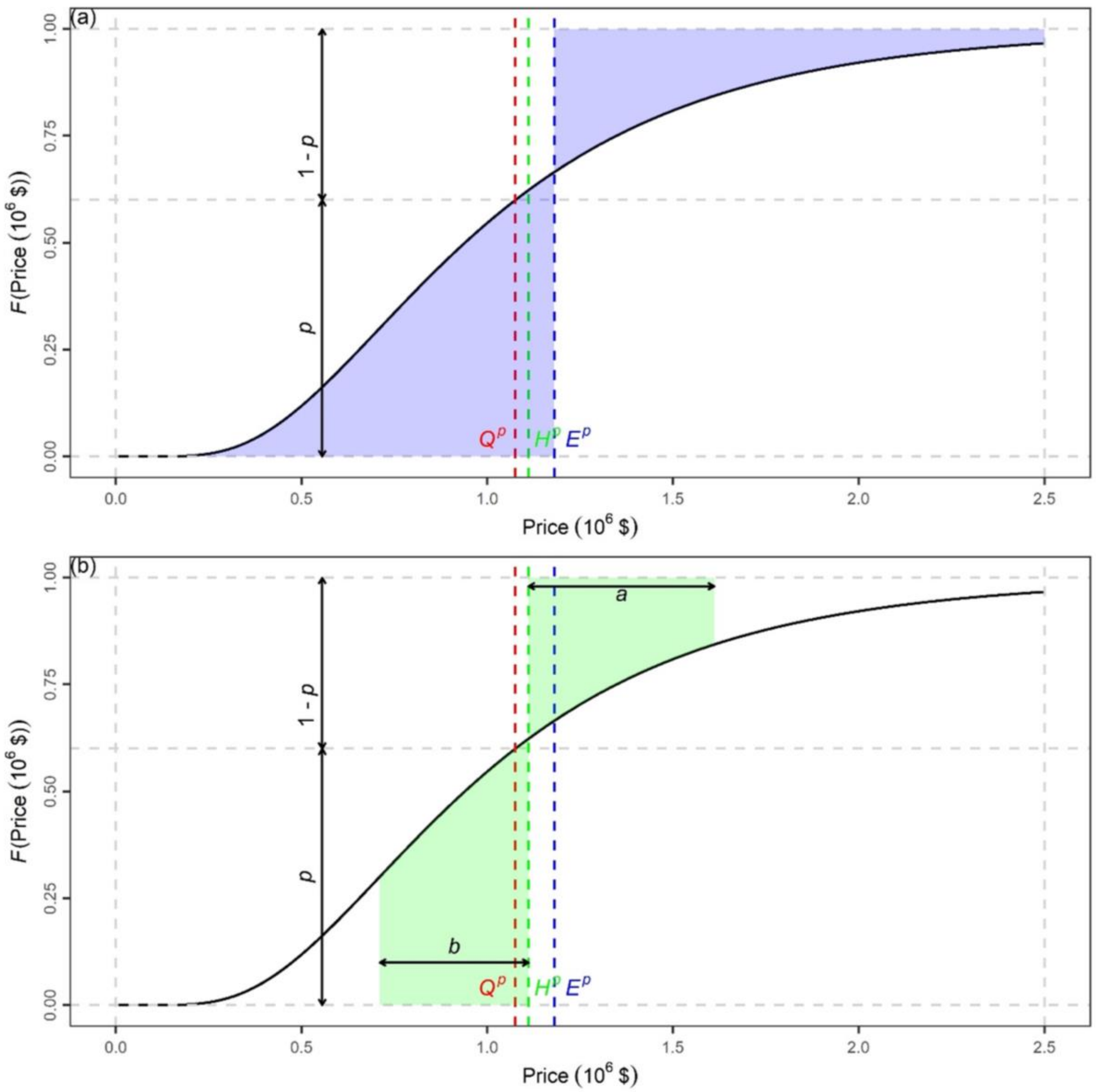

Figure 2. The quantile $Q^p(F)$ (notated with $Q^p$), expectile $E^p(F)$ (notated with $E^p$) and Huber quantile $H^p_{a,b}$ (notated with $H^p$) for the log-normal distribution (with CDF equal to $F$(Price)) with $\mu = -0.063$ and $\sigma = 0.534$, and for level $p = 0.6$, and Huber quantile parameters $a = 0.5, b = 0.4$.

Quantiles, expectiles and Huber quantiles shown in Figure 2 are examined for the level $p = 0.6$, while parameters of the Huber quantile are set equal to $a = 0.5$ and $b = 0.4$. Measurement units for parameters $a$ and $b$ is $10^6$ \$. The vertical line that crosses the price axis (x-axis) at quantile $Q^p(F)$, intersects the CDF curve at a point at which the ratios of lengths of the two vertical segments is $p/(1-p)$. Regarding the expectile $E^p(F)$, the areas of the two shaded regions in Figure 2a are equal to $p/(1-p)$, while for the Huber quantile $H^p_{a,b}$, the areas of the two shaded regions in Figure 2b (bounded to distance $a$ on the right and distance $b$ to the left) are equal to $p/(1-p)$. From the geometric interpretation it is obvious that the family of Huber quantiles $H^\tau_{a,a}(F)$ nests the quantiles $Q^\tau(F)$ (for $a = 0$) and expectiles $E^\tau(F)$ (for $a = \infty$) as edge cases.

## 2.3 Scoring functions

### *2.3.1 Consistency and elicitability*

When making a point prediction, the modeler should receive a directive regarding the procedure to be followed (Murphy and Daan 1985). The perspective developed by Gneiting (2011) identifies two ways to do so. The first way is to disclose the scoring function ex ante to the modeler. The second way is to request from the modeler a specific functional of the predictive distribution; recall from Section 2.2, that quantiles, expectiles and Huber quantiles (as well as their special cases, i.e. the median, the mean and the Huber mean) are functionals of the predictive distribution.

Now let $y$ be an observation and $x$ be the respective prediction. The following definitions refer to the case of dimension $d = 1$, where $x$ and $y$ are real numbers taking values in the interval $I \subseteq \mathbb{R}$. Then a scoring function $S$ is a mapping

$$S: I \times I \to [0, \infty) \tag{16}$$

that assigns a penalty (loss) $S(x, y)$. Among required assumptions, to design a scoring function, an important one is that $S(x, y) \geq 0$, with equality when $x = y$. In the following, scoring functions will be negatively oriented, i.e. the lower the score, the better the prediction. The concept of consistency (Murphy and Daan 1985) for a scoring function is notably important. The following definition of consistency is from Gneiting (2011).

**Definition 1**: The scoring function $S$ is consistent for the functional $T$ relative to the class $\mathcal{F}$ if

$$\mathbb{E}_F[S(t, \underline{y})] \leq \mathbb{E}_F[S(x, \underline{y})] \tag{17}$$

for all probability distributions $F \in \mathcal{F}$, all $t \in T(F)$ and all $x \in I$. It is strictly consistent if it is consistent and equality in eq. (17) implies that $x \in T(F)$.

Returning to the first way of receiving a directive, we assume that the scoring function has been disclosed to the modeler. Then the optimal point prediction of $y$ under the probability distribution $F \in \mathcal{F}$ for the future $\underline{y}$ is (Gneiting 2011):

$$\hat{x} = \arg\min_{x \in I} \mathbb{E}_F[S(x, \underline{y})] \tag{18}$$

Here $\hat{x}$ is a Bayes rule (Ferguson 1967, Robert 2007), and is an optimal decision if one decides to minimize the expected loss.

Assuming that the modeler received a directive to predict a functional, the following theorem by Gneiting (2011) highlights the importance of implementing a consistent scoring function to evaluate the optimality of point predictions.

**Theorem 1**: The scoring function $S$ is consistent for the functional $T$ relative to the class $\mathcal{F}$ if and only if, given any $F \in \mathcal{F}$, any $x \in T(F)$ is an optimal point prediction under $S$.

In other words, if the modeler receives a directive to predict a functional, it is critical that the scoring function for the functional is consistent, i.e. the expected score is minimized when following the directive.

Now we turn to the important concept of elicitability of functionals that first appeared in Osband (1985), while the term elicitable was coined by Lambert et al. (2008). The following definition is from Gneiting (2011) and characterizes the duality of scoring functions and functionals.

**Definition 2**: The functional $T$ is elicitable relative to the class $\mathcal{F}$ if there exists a scoring function $S$ that is strictly consistent for $T$ relative to $\mathcal{F}$.

*2.3.2 Examples of consistent scoring functions*

The concepts of consistency and elicitability become concrete in practical situations. In our case, we assume that the modeler receives a directive to predict the Huber quantile $H_{a,b}^{\tau}(F)$. If the prediction is equal to $x$ and $y$ has materialized, a general class of consistent scoring functions to evaluate the prediction has been defined by Taggart (2022b):

$$S_H(x, y; \tau, \varphi, a, b) = |\mathbb{1}\{x \geq y\} - \tau|(\varphi(y) - \varphi(\kappa_{a,b}(x - y) + y) + \kappa_{a,b}(x - y)\varphi'(x)) \quad (19)$$

where $\varphi$ is a convex function with subgradient $\varphi'$. $S_H$ is strictly consistent for the Huber quantile $H_{a,b}^{\tau}(F)$, if $\varphi$ is strictly convex. Edge cases of the $S_H$ include consistent scoring functions for the quantile and expectile functionals.

The general class of consistent scoring functions for the quantile functional $Q^{\tau}(F)$ is defined by (Thomson 1979, Saerens 2000):

$$S_Q(x, y; \tau, g) = |\mathbb{1}\{x \geq y\} - \tau||g(x) - g(y)| \quad (20)$$

where $g$ is a nondecreasing function. The generalized piecewise linear scoring function of order $\tau \in (0, 1)$ $S_Q$ is strictly consistent for the quantile $Q^{\tau}(F)$, if $g$ is strictly increasing.

The general class of consistent scoring functions for the expectile functional $E^{\tau}(F)$ is defined by (Gneiting 2011):

$$S_E(x, y; \tau, \varphi) = |\mathbb{1}\{x \geq y\} - \tau|(\varphi(y) - \varphi(x) + \varphi'(x)(x - y)) \quad (21)$$

where $\varphi$ is a convex function with subgradient $\varphi'$. $S_E$ is strictly consistent for the expectile $E^\tau(F)$, if $\varphi$ is strictly convex.

The following relationships between $S_H$ and its edge cases $S_Q$, $S_E$ are from Taggart (2022b), while their proof is straightforward:

$$\lim_{a\to\infty} S_H(x, y; \tau, \varphi, a, a) = S_E(x, y; \tau, \varphi) \quad (22)$$

$$\lim_{a\downarrow 0} S_H(x, y; \tau, \varphi, a, a)/a = S_Q(x, y; \tau, \varphi') \quad (23)$$

In the following we will work with the choice of the strictly convex $\varphi(t) = t^2$. Then, from eq. (19), the generalized Huber scoring function $S_{a,b}(x, y; \tau)$ arises (Taggart 2022b):

$$S_{a,b}(x, y; \tau) = |\mathbb{1}\{x \geq y\} - \tau|(y^2 - (\kappa_{a,b}(x - y) + y)^2 + 2x\kappa_{a,b}(x - y)) \quad (24)$$

Combining eqs. (22) and (23) with eq. (24), results in the following respective edge cases of $S_{a,b}(x, y; \tau)$:

$$S_1(x, y; \tau) = 2|\mathbb{1}\{x \geq y\} - \tau||x - y| \quad (25)$$

$$S_2(x, y; \tau) = |\mathbb{1}\{x \geq y\} - \tau|(x - y)^2 \quad (26)$$

$S_1(x, y; \tau)$ is up to a multiplicative constant the asymmetric piecewise linear scoring function (also termed as tick-loss or quantile loss in the literature) which is strictly consistent for the quantile $Q^\tau(F)$ (Raiffa and Schlaifer 1961), and lies at the heart quantile regression (Koenker and Bassett Jr 1978). $S_2(x, y; \tau)$ is strictly consistent for the expectile $E^\tau(F)$.

Setting $a = b \geq 0$, and $\tau = 1/2$ in eq. (24), results in

$$L_a(x, y) = \begin{cases} (x - y)^2/2, |x - y| \leq a \\ a|x - y| - a^2/2, |x - y| > a \end{cases} \quad (27)$$

The scoring function $L_a(x, y)$ has been proposed by Huber (1964) (and coined as Huber loss in the literature) and lies at the heart of Huber regression. The scoring function $L_a(x, y)$ is strictly consistent for the mean Huber functional $H_{a,a}^{1/2}(F)$ (recall its definition from Section 2.2.2). Combining eqs. (22) and (23) with eq. (27), results in the following edge cases of $L_a(x, y)$ respectively:

$$L_1(x, y) = |x - y| \quad (28)$$

$$L_2(x, y) = (x - y)^2/2 \quad (29)$$

$L_1(x, y)$ is the absolute error (AE) scoring function and $L_2(x, y)$ is equal to the squared

error (SE) scoring function up to a multiplicative constant. The AE and the SE scoring functions are strictly consistent for the median and mean functionals respectively. Another way to reach $L_1(x, y)$ and $L_2(x, y)$ is to set $\tau = 1/2$ in $S_1(x, y; \tau)$ and $S_2(x, y; \tau)$ respectively.

Notably the class of scoring functions

$$S_E(x, y; 1/2, \varphi) = (1/2)(\varphi(y) - \varphi(x) + \varphi'(x)(x - y)) \tag{30}$$

that results from eq. (21) for $\tau = 1/2$, is consistent for the mean functional (Savage 1971). Functions of the form $S_E(x, y; 1/2, \varphi)$ up to a multiplicative constant are referred as Bregman functions by Banerjee et al. (2005).

The existence of strictly consistent scoring functions for the quantiles, expectiles and Huber quantiles implies that they are elicitable (recall Definition 2 of elicitability).

*2.3.3 Average and skill scores*

In practical situations, one has a test set with sample size $n$, predictions $x_i, i = 1, \dots, n$ of a functional and respective materializations $y_i, i = 1, \dots, n$ of the random variable $\underline{y}$. Then the summary measure of the predictive performance takes the form of the (negatively oriented) average score $\bar{S}$

$$\bar{S} = (1/n) \sum_{i=1}^{n} S(x_i, y_i) \tag{31}$$

When comparing two methods, one is said to outperform the other, when its average score is smaller. In some cases, one has to compare multiple methods at multiple datasets with data that take values ranging at different magnitudes. That is a common case, when comparing the performance of multiple forecasting methods at multiple time series. Then skill scores can be used to assess multiple methods. A skill score takes the form (Gneiting 2011):

$$\bar{S}_{\text{skill}} := 1 - \bar{S}_{\text{method}} / \bar{S}_{\text{ref}} \tag{32}$$

$\bar{S}_{\text{ref}}$ is the average score of the reference method (frequently the simpler method among the competing ones) and $\bar{S}_{\text{method}}$ is the average score of the method of interest. The skill score $\bar{S}_{\text{skill}} \leq 1$. When $\bar{S}_{\text{method}} = 0$ (which corresponds to a perfect prediction from the method of interest), then $\bar{S}_{\text{skill}} = 1$, while when $\bar{S}_{\text{method}} = \bar{S}_{\text{ref}}$, then $\bar{S}_{\text{skill}} = 0$ and both methods have equal predictive performance. When the method of interest outperforms the reference method, we have $\bar{S}_{\text{method}} < \bar{S}_{\text{ref}}$, that implies $\bar{S}_{\text{skill}} > 0$, while when the reference method outperforms the method of interest we have $\bar{S}_{\text{skill}} < 0$. In general, the higher the skill score, the better the performance of the methods of interest compared to the reference method. The concepts of consistency and elicitability continue to apply for skill scores at least when the test sample size $n$ is large (Gneiting 2011).

## 2.4 Economic interpretation

### 2.4.1 Mixture representations

Quantile functionals are rather straightforward to interpret. Furthermore, scientists and practitioners are familiar with them given that they are literally integral parts of probabilistic predictions. The interpretability of expectiles has been questioned (Kneib 2013), while for Huber quantiles, the situation is even worse given the inclusion of the two additional parameters $a$ and $b$. A natural economic interpretation of expectiles (and respective scoring functions) has been introduced by Ehm et al. (2016) and has been extended to the case of Huber quantile functionals by Taggart (2022b). The interpretation is associated to a binary investment decision with a random outcome $y$ (Ehm et al. 2016) and is based on mixture representations of quantile, expectile and Huber quantile scoring functions.

The following results on the mixture representation of the quantile scoring function family $S_Q(x, y; \tau, g)$ and the expectile scoring function family $S_E(x, y; \tau, \varphi)$ are due to Ehm et al. (2016), while the mixture representation for the Huber quantile scoring function family $S_H(x, y; \tau, \varphi, a, b)$ is due to Taggart (2022b).

The mixture representation of $S_Q(x, y; \tau, g)$ is

$$S_Q(x, y; \tau, g) = \int_{-\infty}^{\infty} S_{\tau,\theta}^{Q}(x, y) \, \mathrm{d}M(\theta) \, , (x, y) \in \mathbb{R}^2 \tag{33}$$

where

$$S_{\tau,\theta}^{Q}(x, y) = \begin{cases} 1 - \tau, y \leq \theta < x \\ \tau, x \leq \theta < y \\ 0, \text{otherwise} \end{cases} \tag{34}$$

and $M$ is non-negative measure. The mixing measure $M$ is unique and satisfies $\mathrm{d}M(\theta) = \mathrm{d}g(\theta)$ where $\theta \in \mathbb{R}$ and $g$ is the nondecreasing function in the representation of eq. (20).

The mixture representation of $S_E(x, y; \tau, \varphi)$ is

$$S_E(x, y; \tau, \varphi) = \int_{-\infty}^{\infty} S_{\tau,\theta}^{E}(x, y) \, \mathrm{d}M(\theta) \, , (x, y) \in \mathbb{R}^2 \tag{35}$$

where

$$S^{E}_{\tau,\theta}(x,y) = \begin{cases} (1-\tau)|\theta - y|, y \leq \theta < x \\ \tau|\theta - y|, x \leq \theta < y \\ 0, \text{otherwise} \end{cases} \quad (36)$$

and $M$ is non-negative measure. The mixing measure $M$ is unique and satisfies $\mathrm{d}M(\theta) = \mathrm{d}\varphi'(\theta)$ where $\theta \in \mathbb{R}$ and $\varphi'$ is the left-hand derivative of the convex function $\varphi$ in the representation of eq. (21).

The mixture representation of $S_H(x,y;\tau,\varphi,a,b)$ is

$$S_H(x,y;\tau,\varphi,a,b) = \int_{-\infty}^{\infty} S^{H}_{\tau,a,b,\theta}(x,y)\,\mathrm{d}M(\theta)\,, (x,y) \in \mathbb{R}^2 \quad (37)$$

where

$$S^{H}_{\tau,a,b,\theta}(x,y) = \begin{cases} (1-\tau)\min\{\theta - y, b\}, y \leq \theta < x \\ \tau \min\{y-\theta, a\}, x \leq \theta < y \\ 0, \text{otherwise} \end{cases} \quad (38)$$

and $M$ is non-negative measure. The mixing measure $M$ is unique and satisfies $\mathrm{d}M(\theta) = \mathrm{d}\varphi'(\theta)$ where $\theta \in \mathbb{R}$ and $\varphi'$ is the left-hand derivative of the convex function $\varphi$ in the representation of eq. (19).

*2.4.2 House investment decision*

The elementary scoring function $S^{H}_{\tau,a,b,\theta}(x,y)$ for the Huber quantile functional admits an economic interpretation. Here we adapt the example in Section 5.3 of Taggart (2022b) to the case of real estate investment. We assume that one considers investing an amount $\theta$ to buy a house that later can be sold with an unknown amount $y$. The real price of the house is not known, but one may predict it using covariates (e.g. area of the house, year built and sell prices). Profit from the investment is possible if and only if $y > \theta$. We assume that there is a limit (cap) $b$ on the losses one can incur (perhaps due to limited budget) and a limit (cap) $a$ on profits one can receive (perhaps due to some legal agreements). The pay-off structure for the $S^{H}_{\tau,a,b,\theta}(x,y)$ elementary scoring function is summarized in the upper half of Table 1 and can be described as follows:

- If the investor refrains from the deal, the payoff is 0.
- If one decides to invest and $y \leq \theta$, realizes, then the payout is $-(1-r_L)\min\{\theta - y, b\} \leq 0$. Here the loss $\theta - y$ is bounded by $b$, while $r_L \in [0,1)$ is the deduction rate and $1 - r_L$ accounts for reduction in income tax.
- If one decides to invest and $y > \theta$, realizes, then the pay-off (gain) is $(1 - r_G)\min\{y - \theta, a\} > 0$, where $r_G \in [0,1)$ is the tax rate applied to the profits.

Table 1. Overview of pay-off structure for decision rule to invest if and only if $x > \theta$.

| | $y \leq \theta$ | $y > \theta$ |
|---|---|---|
| Monetary payoff | | |
| $x \leq \theta$ | 0 | 0 |
| $x > \theta$ | $-(1 - r_L)\min\{\theta - y, b\}$ | $(1 - r_G)\min\{y - \theta, a\}$ |
| Score (regret) | | |
| $x \leq \theta$ | 0 | $(1 - r_G)\min\{y - \theta, a\}$ |
| $x > \theta$ | $(1 - r_L)\min\{\theta - y, b\}$ | 0 |

Assuming that an investor can issue perfect predictions for the house price, it is straightforward to reformulate the upper half of Table 1 to the lower half that is a regret matrix. The regret matrix is negatively oriented; therefore if one is willing to minimize loss, he should invest if and only if $x > \theta$, where $x$ is the predictive Huber quantile $H^{\tau}_{a,b}(F)$, where $\tau = (1 - r_G)/(2 - r_L - r_G)$.

When $r_L = r_G$, then $\tau = 1/2$ (case of the mean Huber quantiles), while when $a = \infty$ and $b = \infty$, then we assume there is no cap on profits or losses and we have the case of economic interpretation of expectiles.

## 3. Deep learning algorithms

An integral part of deep learners is the objective function (Lecun et al. 2015, Schmidhuber 2015). The whole idea of DHQRN is based on the adaptation of the objective function to estimate the DL algorithm and subsequently predict Huber quantiles (see Section 3.1). Further details of the deep learners implemented are described in Section 3.2 (model architectures) and Section 3.3 (preprocessing and optimization).

### 3.1 Estimation – setting the objective function

Technically, training a DL model with a Huber quantile scoring function is a conditional Huber quantile estimation problem in the sense described by Koenker and Bassett (1978) for the case of a linear quantile regression framework and by Dimitriadis et al. (2024) for the general case of regression models. Thus, the problem of training the DL model is similar to regular cases, in which the objective (loss) function is not of a conventional type (e.g., an absolute or a squared error scoring function). Therefore, the procedure is applicable to any architecture; see Figure 3. When new observations of predictor variables arrive, then the trained model will predict conditional Huber quantiles. The following procedures (setting DL architectures, preprocessing and optimization) do not differ from the current state-of-the-art procedures of a regular DL model.

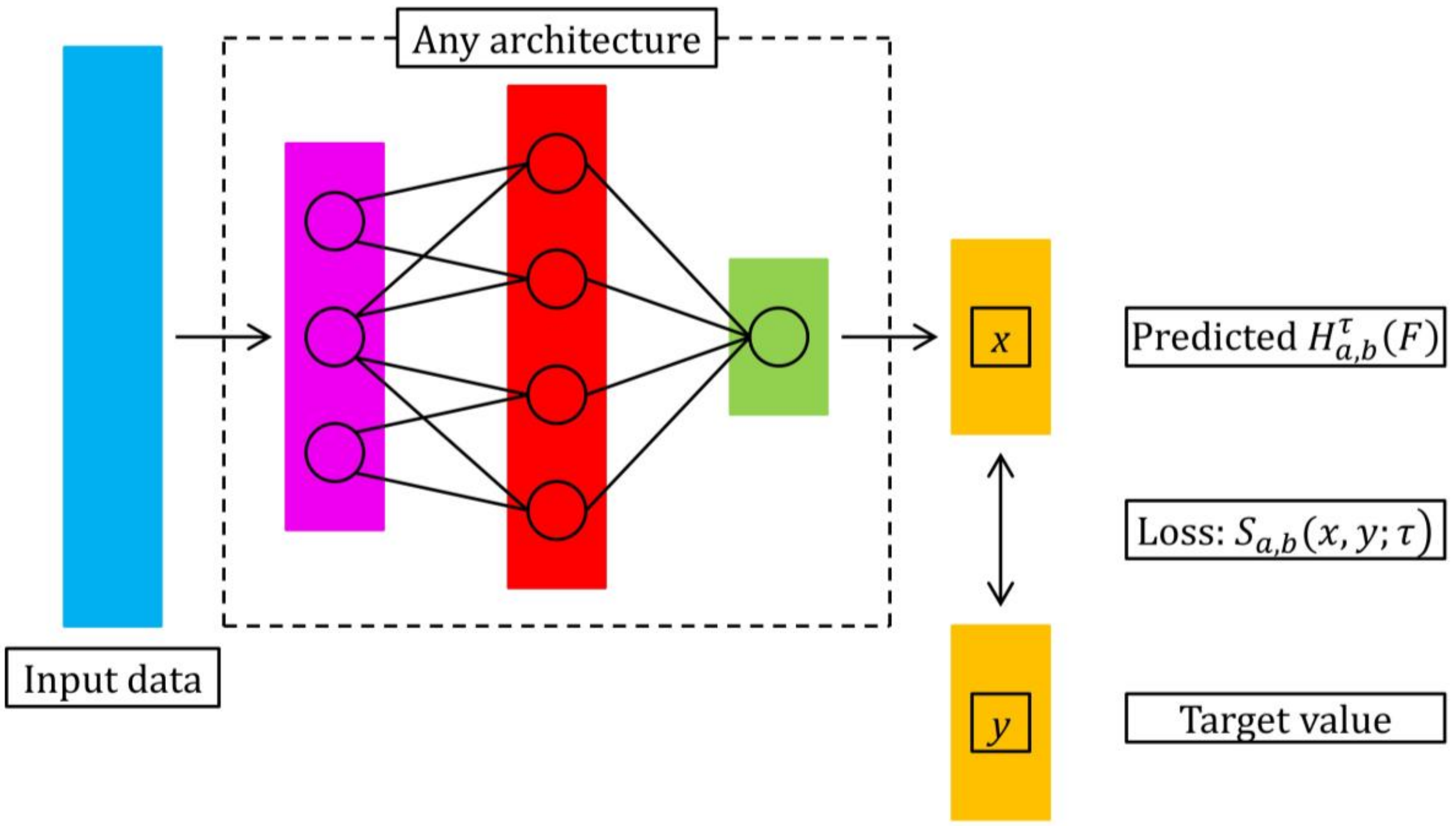

Figure 3. General architecture of DHQRN.

### 3.2 Architectures

We compared the following three DL architectures developed with the contributed `Python` library `tensorflow` v.2.12.0 (Abadi et al. 2016a, 2016b, TensorFlow Developers 2023):

- Model #1: Three regular densely-connected neural networks layers (hereinafter called dense layers) with 64 units (units refers to the dimensionality of the output space of the layer) followed by one dense layer with 32 units.

- Model #2: Two dense layers with 64 units followed by a dropout layer with dropout rate equal to 0.5 and a dense layer with 32 units.

- Model #3: Two dense layers with 64 units.

Inputs for all models include 11 predictor variables for the case of Melbourne houses and 12 predictor variables for the case of Boston houses (see Section 4 for data descriptions of the two applications) and the output is a single variable (single value). In subsequent applications, we keep the three architectures as-is, while we adapt their objective function. We could also assess other architectures, perhaps using some spatial features, and predict more accurate Huber quantiles, yet this is out of the scope of the study. Yet the herein proposed architectures suffice for the demonstration of DHQRN.

### 3.3 Preprocessing and optimization

Predictor variables were normalized in the usual way, while we were attentive, when necessary, to exclude information from sets used for validation and testing. In particular, the primary dataset (that is described in more detail in Section 4.1) includes 8 843 houses prices and respective predictor observations in Melbourne. We split the dataset randomly to three sets, i.e. a training set with 3 537 houses (40% of the samples), a validation set with 2 653 houses (30% of the samples) and a test set with 2 653 houses (30% of the samples). Therefore, we normalized the predictors in the training set while we also normalized the predictors in the training-validation set (that includes 6 190 houses of the training and validation set). A similar setting was followed for the supporting application of Boston house price prediction (see Section 4.3).

For training each model we selected rectified linear unit (ReLU) activation functions (Schmidt-Hieber 2020), the ADAM (Adaptive Moment Estimation, a variant of stochastic gradient descent) optimizer with ADAM parameter equal to 0.005 (Kingma and Ba 2015), early stopping with patience equal to 10 (Prechelt 1998), dropout (only for Model #2) with dropout rate equal to 0.5 (Srivastava et al. 2014), 200 epochs for training and batch size equal to 32 (Smith et al. 2018).

More specifically, we trained a model in the normalized training set and performed early stopping to estimate the optimal number of epochs in the non-normalized validation set. Then we estimated the parameters of the model for the optimal number of epochs in the normalized training-validation set and used the non-normalized test set for assessing the algorithms.

Since we set the Huber quantile scoring function as the objective function, the assessment of the models is based on the skill scores derived from the same scoring function (see Section 2.3.3), where a baseline model that uses sample Huber quantiles of the marginal distribution of the test set is used as a reference method.

While further hyperparameter tuning, like applying different activation functions, adjusting batch size, exploring various optimizers and dropout options, could potentially improve performance, it falls outside the scope of this paper. Regularization techniques might also be beneficial. These explorations are left for future studies. Prediction competitions could be a valuable tool in this direction.

## 4. Application

The three models introduced in Section 3.2 are trained to predict house prices in Melbourne, Australia. In Sections 4.1 and 4.2, the dataset and the results from the implementation of the algorithms are presented. A supporting application to house price predictions in Boston with the same DL architectures is presented in Section 4.3. Finally a simulation experiment in Section 4.4 (supported by code provided as supplementary information), allows assessing stylized versions of the algorithm.

### 4.1 Data

#### *4.1.1 Data description*

The Melbourne house prices dataset is available at https://www.kaggle.com/datasets/anthonypino/melbourne-housing-market (date of access: 2023-04-02). The dataset includes data for 34 857 houses in Melbourne. We worked in a reduced dataset with 8 843 houses, excluding those houses with incomplete data (e.g., due to missing values in the predictors). The locations of the 8 843 houses used for modelling are shown in Figure 4.

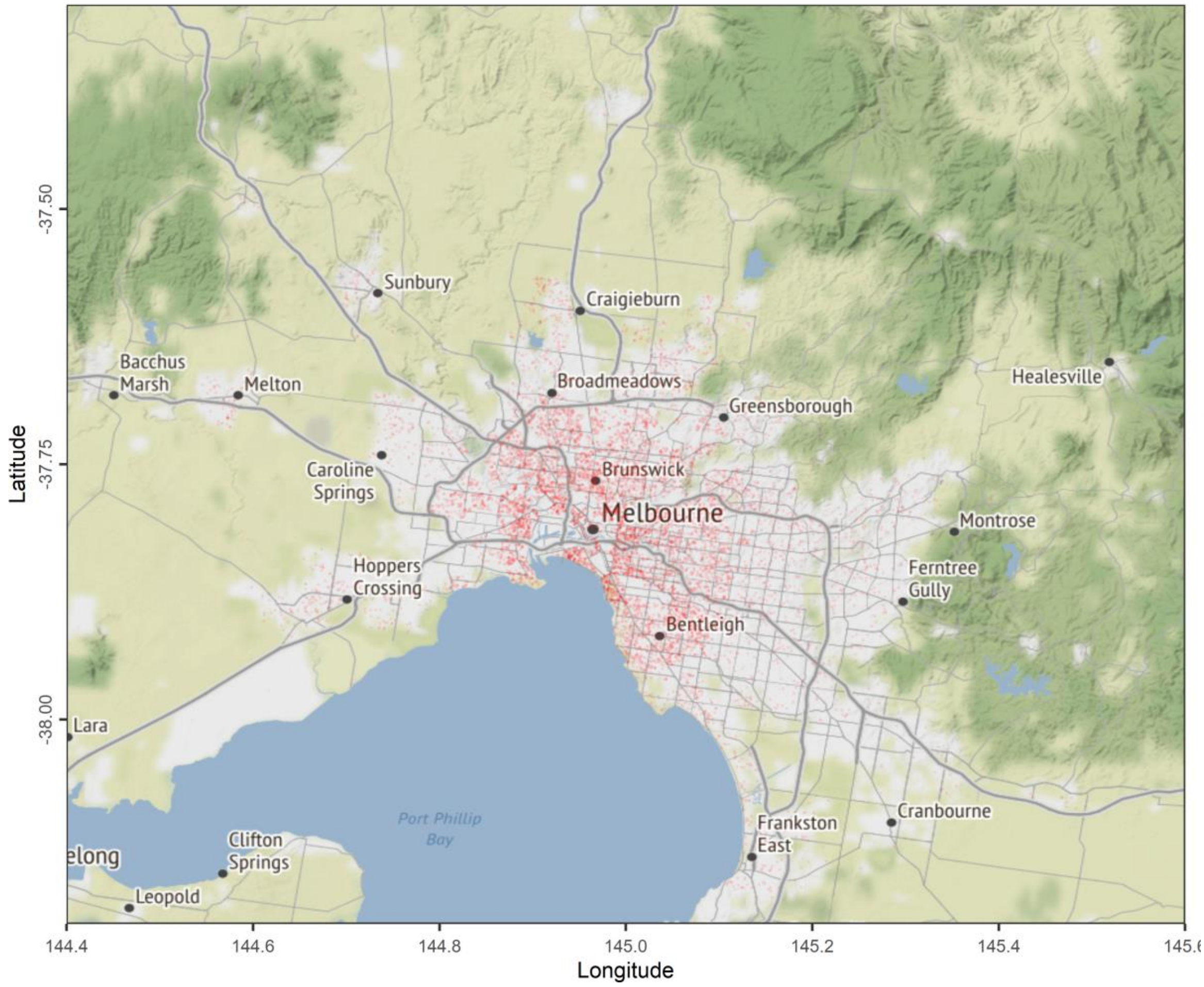

Figure 4. Locations of houses in Melbourne.

In the regression problem, the predictor variables for the house price are the building area, land size, year built and year sold, number of rooms, number of bedrooms, number of carspots and distance from CDB (Melbourne central business district). Value ranges for the predictors in the full dataset are shown in Figure 5.

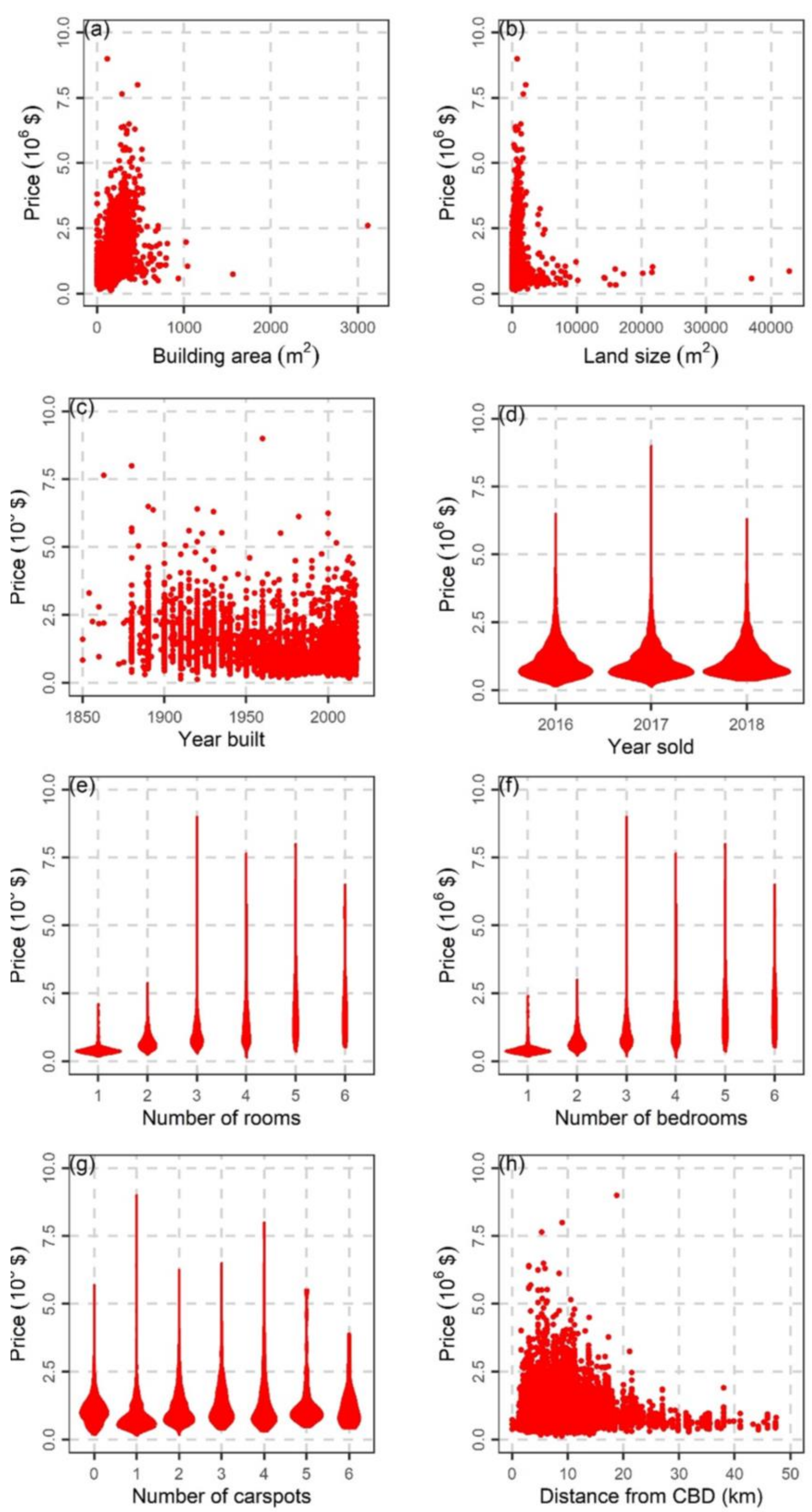

Figure 5. Visualization of the relationship between house prices in Melbourne with (a) building area, (b) land size, (c) year built, (d) year sold, (e) number of rooms, (f) number of bedrooms, (g) number of carspots, (h) distance from CBD (Melbourne central business district).

A histogram of the house prices is shown in Figure 6. The log-normal probability distribution is a frequent choice for modelling house prices (Razen et al. 2023). We fitted a log-normal probability distribution (with probability density function defined by eq. (15)), using the exact maximum likelihood estimator (implemented by the `fitdistr` function of the `MASS` `R` package, Venables and Ripley 2002). The log-normal probability distribution fits well the house prices as evident in Figure 6. The respective cumulative distribution function is presented in Figure 7. Please recall from Figure 2, that we used the same distribution (with identical parameters) to graphically present cases of quantiles, expectiles and Huber quantiles.

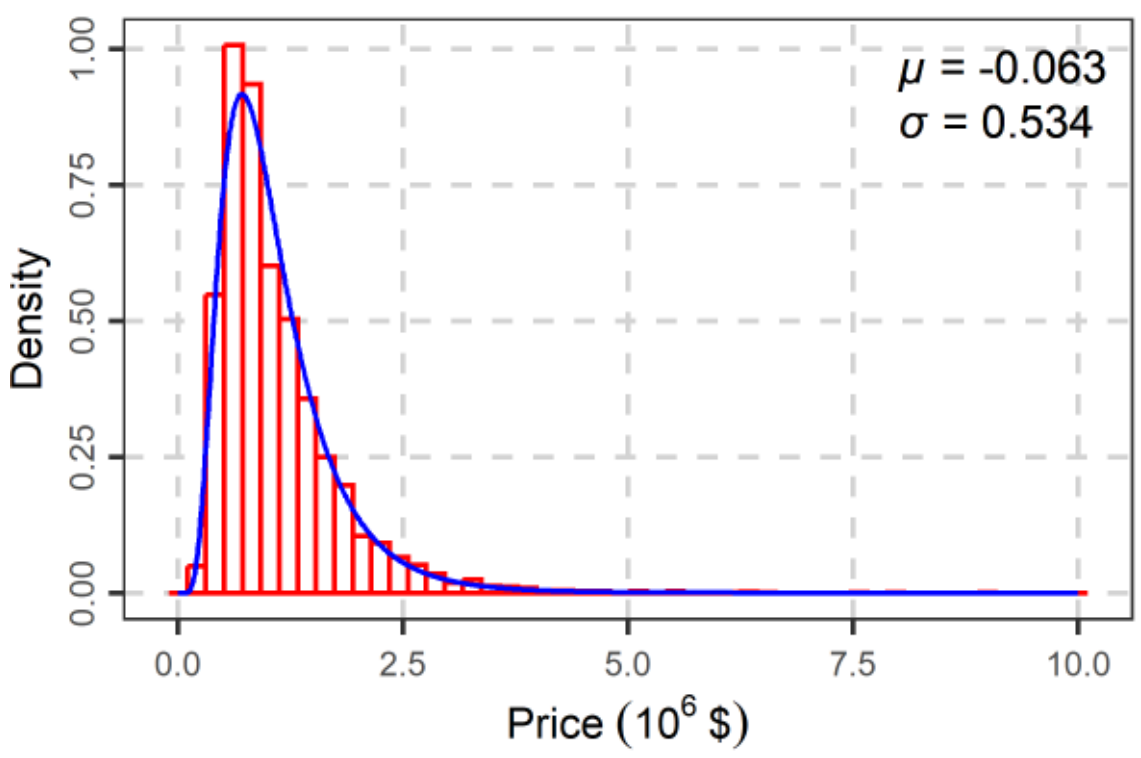

Figure 6. Histogram of house prices (in red) and fitted log-normal distribution (in blue) to house prices. Maximum likelihood estimates of $\mu$ and $\sigma$ are also presented.

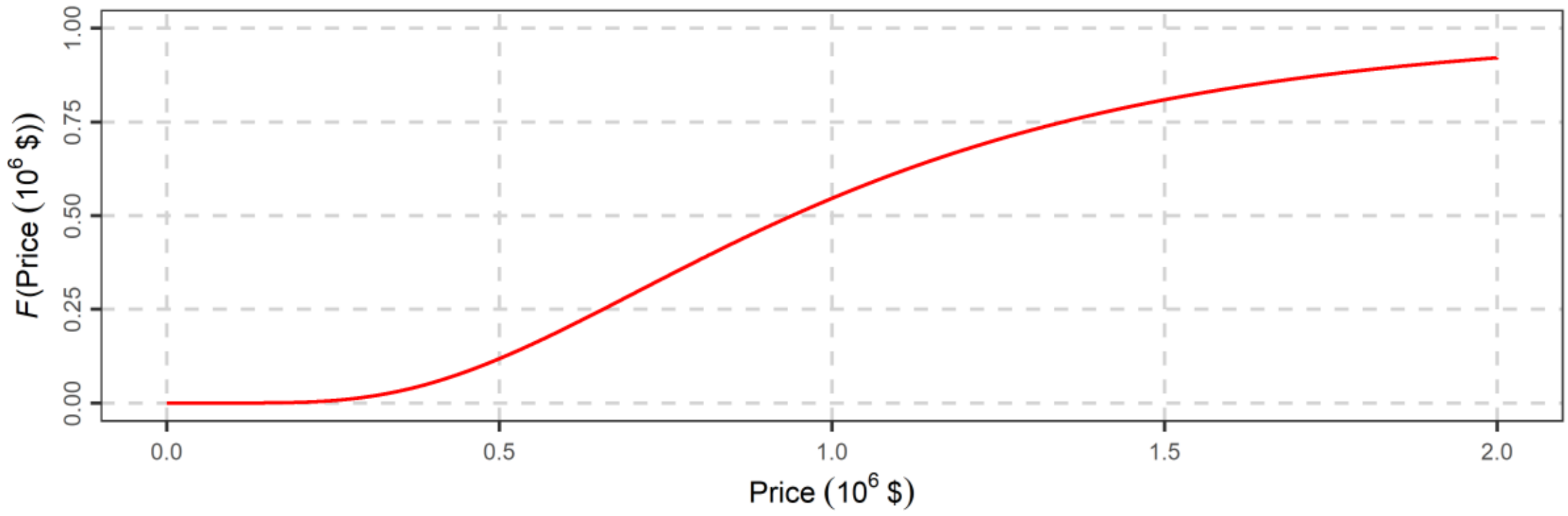

Figure 7. Cumulative distribution function of the fitted log-normal distribution.

## 4.2 Application

The three DL models introduced in Section 3.2 are trained using Huber quantile scoring functions with various parameter sets to predict house prices in the test set according to the procedure described in Section 3. Recall from Section 2.3 that the various sets of parameters of the Huber quantile scoring function correspond to a pure Huber quantile scoring function as well as its edge cases, i.e. the quantile and expectile scoring functions. In the following, comparison results refer to the test set.

*4.2.1 Results*

We trained each model with Huber quantile scoring functions $S_{a,b}(x, y; \tau)$ at levels $\tau \in \{0.4, 0.5, 0.6, 0.7, 0.8\}$, while the remaining parameters $a$ and be $b$ were set equal to 0.5 and 0.4 respectively. Recall from Section 2.4.2, that the values of the parameters $a$ and $b$ correspond to caps on the profits and losses respectively. Furthermore, recall from Section 2.4.2, that each value of $\tau$ corresponds to a specific combination of the income deduction rate $r_L$ and the profit tax rate $r_G$. Obviously, it would be natural to start from setting the deduction rate and the tax rate based on local legislation and then compute the level $\tau$ of interest, but since our example is intended to illustrate general aspects of DHQRN, we proceeded with pre-specified values of levels $\tau$.

To be more specific we compare the three models at level $\tau = 0.4$. The first column of Figure 8 presents the skill score $\bar{S}_{a,b,\tau,\text{skill}}$ for $\tau = 0.4$ and the prespecified values of $a$ and $b$, where $\bar{S}_{a,b,\tau,\text{model } i}$ corresponds to the average score of Model #$i$, for the case of scoring Huber quantile scoring functions $S_{a,b}(x, y; \tau)$ defined by eq. (24).

$$\bar{S}_{a,b,\tau,\text{skill}} := 1 - \bar{S}_{a,b,\tau,\text{model } i} / \bar{S}_{a,b,\tau,\text{mHq}}, i = 1,2,3 \tag{39}$$

Here the reference model is the marginal Huber quantile (mHq), that corresponds to the test set sample Huber quantile at the respective level $\tau$ and values $a = 0.5$ and $b = 0.4$. In particular, the marginal Huber quantile can be computed by the sample equivalent of eq. (11), where the sample is the test set. Marginal quantities of the test set are common baseline models in forecasting applications (Gneiting et al. 2023). It is evident that Model #3 outperforms Models #1 and #2, at level $\tau = 0.4$, since it has the highest skill score. Therefore, to decide whether to invest in buying a house, one should prefer to base the decision on predictions issued by Model #3. Similarly, at remaining quantile levels, Model #3 seems to outperform the other models due to higher skill scores, except of the cases of $\tau = 0.6$ and 0.7, where Model #1 is equal in performance.

It is a common practice to assess multiple algorithms designed for different tasks, when given a directive to predict a functional. For instance, one could request to compare QRNN and DHQRN, in this example. Recall from Section 2.3, that since we were given a directive to predict Huber quantile functionals, it is appropriate to use Huber quantile scoring functions to assess the predictions, while quantile scoring functions would not be consistent for the task of assessment. Therefore, it is not appropriate to compare QRNN and DHQRN in our example, yet QRNN is a special case of DHQRN.

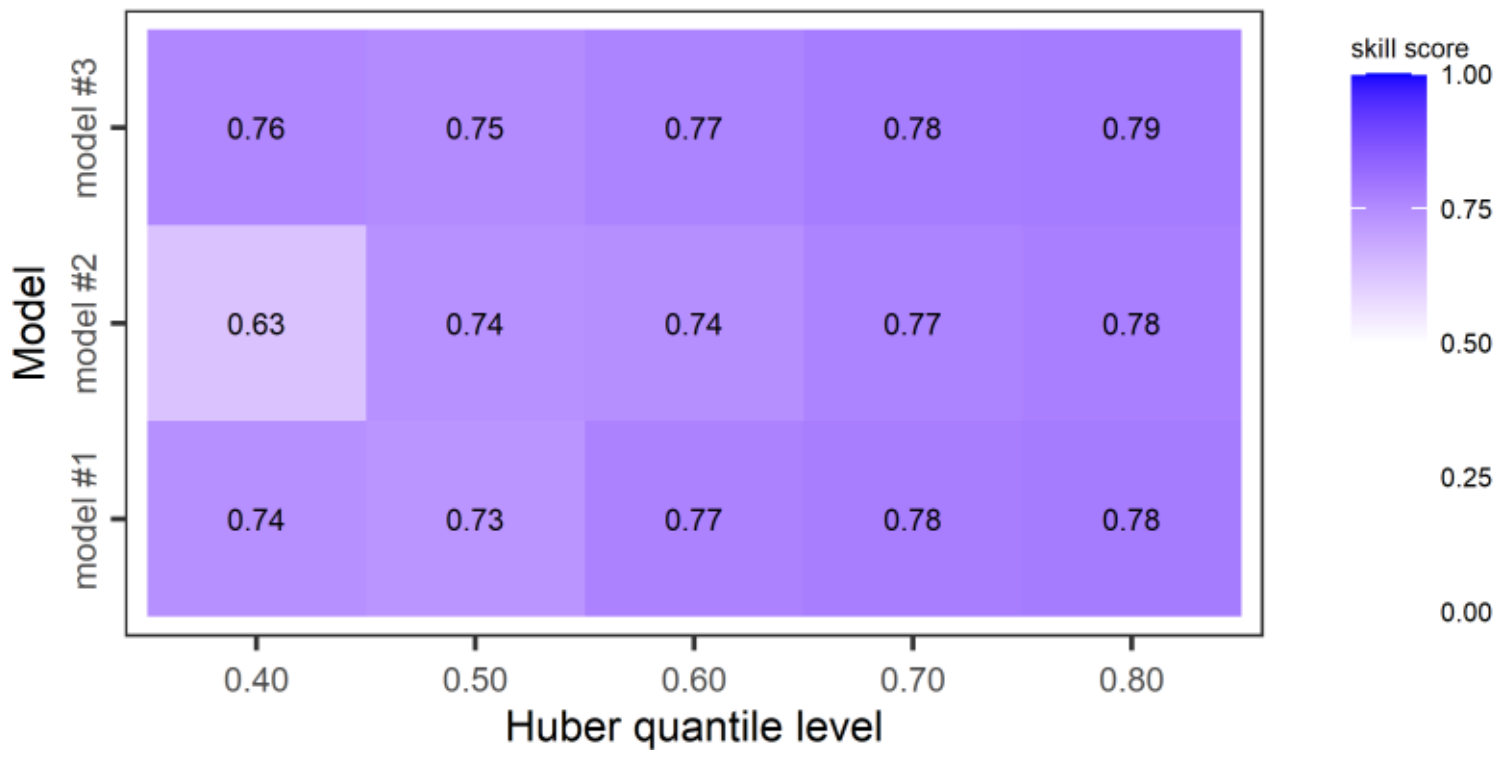

Figure 8. Skill scores for the Huber quantile scoring function $S_{a,b}(x, y; \tau)$ for Models #1, #2 and #3 using the marginal Huber quantile as a reference method, at Huber quantile levels $\tau \in \{0.4, 0.5, 0.6, 0.7, 0.8\}$ with parameters of the Huber quantile scoring function $a = 0.5$ and $b = 0.4$.

Skill scores report relative performances between models, hence they have limited interpretability. Based on predictions on the test set one can estimate the sample equivalent of the respective level from eq. (14). Those sample estimates are presented in Figure 9, which correspond to predictions from the trained models of Figure 8. Regarding Model #1 and the case of the nominal Huber quantile level 0.4 (first column of Figure 9) the predicted Huber quantile level is equal to 0.36. If the predictions of Model #1 were perfect, then the sample estimate of the quantile level should be equal to the nominal level $(= 0.4)$. A model with Huber quantile sample estimate closer to the nominal level should be preferable. For instance, for the case of nominal Huber quantile level 0.4, Model #3 outperforms the other models, because its sample estimate is equal to 0.43, which corresponds to the minimum distance $0.03(= |0.40 - 0.43|)$ in the first column. In general, the model assessment based on Figure 9 is consistent with that of Figure 8, although we would prefer to use skill scores due to their favourable properties (consistency) as explained in Section 2.3. Please also recall from Section 2.2.2 that Huber quantile frequencies are the equivalent of frequencies for quantiles. In particular, frequencies for quantiles can be estimated by eq. (12).

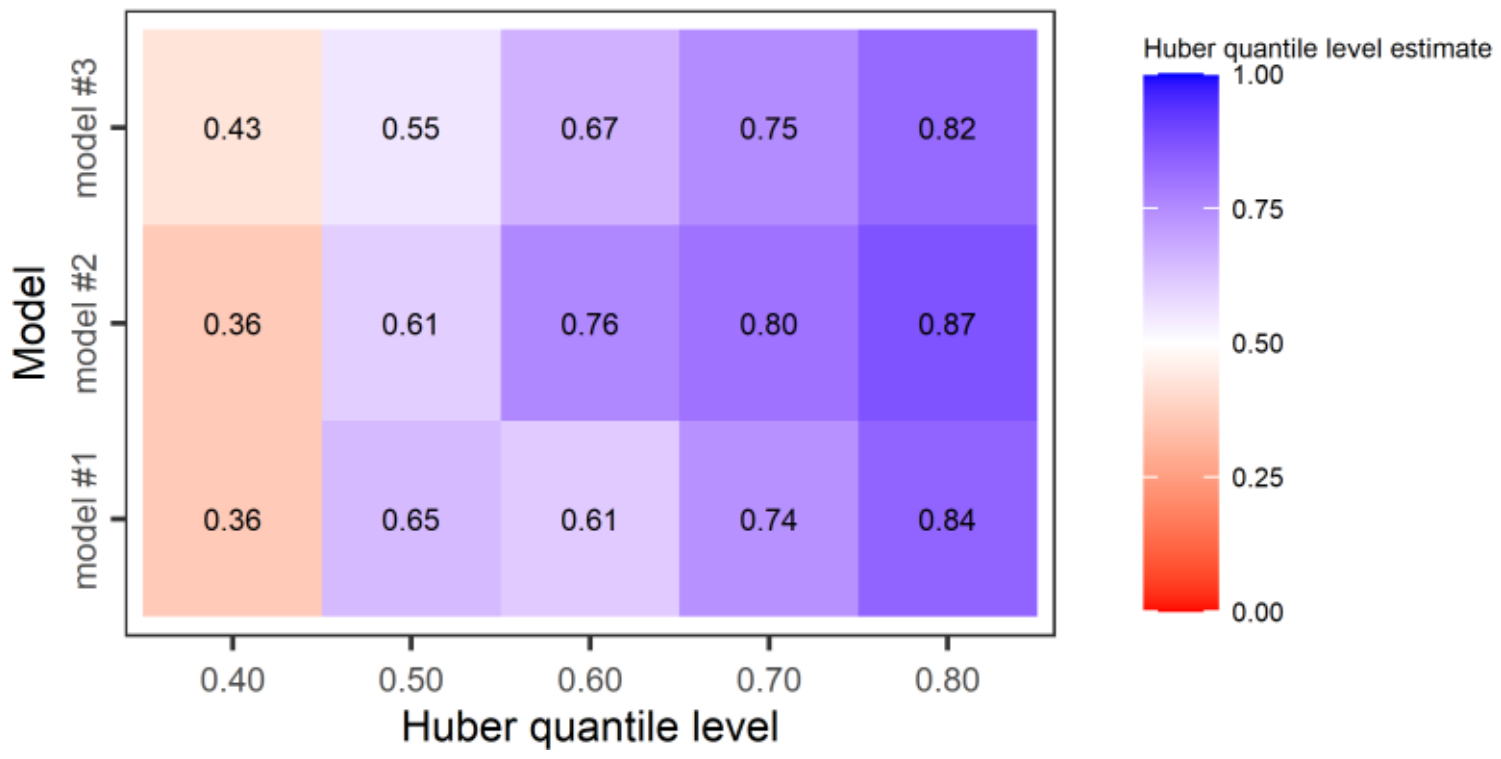

Figure 9. Huber quantile level sample estimates for Models #1, #2, and #3 trained with the Huber quantile scoring function $S_{a,b}(x, y; \tau)$ at Huber quantile levels $\tau \in \{0.4, 0.5, 0.6, 0.7, 0.8\}$ with parameters of the Huber quantile scoring function $a = 0.5$ and $b = 0.4$.

Finally, it is also interesting to know the relation between predicted quantiles, expectiles and Huber quantiles at a specific level. As already discussed in Section 2.2.2, the relationship between quantiles and expectiles depends on the probability distribution and may be explicit in some cases (Jones 1994). Explicit theoretical relations between Huber quantiles and other functionals have not yet been discovered. In Figure 2, we have already shown the relationship between the three functionals for the case of the specific log-normal probability distribution, for specific level and Huber quantile parameters. Based on the geometric interpretation in Section 2.2.2, one would expect that the Huber quantile is located between the quantile and the expectile, while the relative ordering (e.g., when quantiles are smaller compared to Huber quantiles) should depend on the modelled probability distribution.

The relationships between the predicted quantiles, expectiles and Huber quantiles at level $\tau = 0.5$ and all combinations of the parameters $a \in \{0.5, 1, 1.5, 2, 2.5, 3\}$ and $b \in \{0.5, 1, 1.5, 2\}$ of the Huber quantile are shown in Figures 10 and 11. In general, Huber quantiles are higher than quantiles and lower than expectiles, although that is not always the case perhaps due to the Model's inaccurate predictions. As also expected, when parameters $a, b$ are increasing, the respective predicted Huber quantiles move from quantiles to expectiles. This is because expectiles are edge cases of Huber quantiles when $a = \infty$ and $b = \infty$.

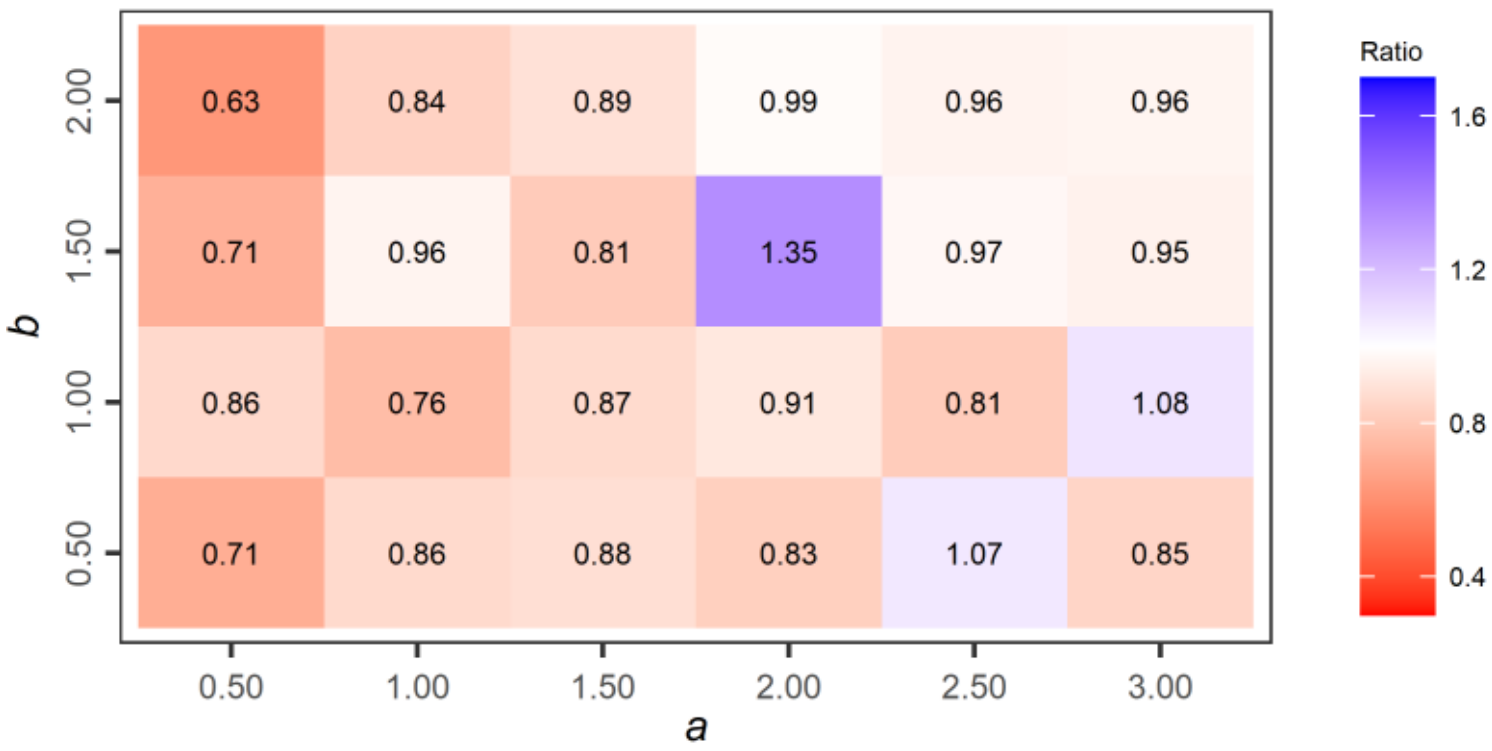

Figure 10. Average ratios $H_{a,b}^{\tau}(F)/E^{\tau}(F)$ of predicted Huber quantiles and predicted expectiles at level $\tau = 0.5$ for all combinations of the parameters $a \in \{0.5, 1, 1.5, 2, 2.5, 3\}$ and $b \in \{0.5, 1, 1.5, 2\}$ of the Huber quantile.

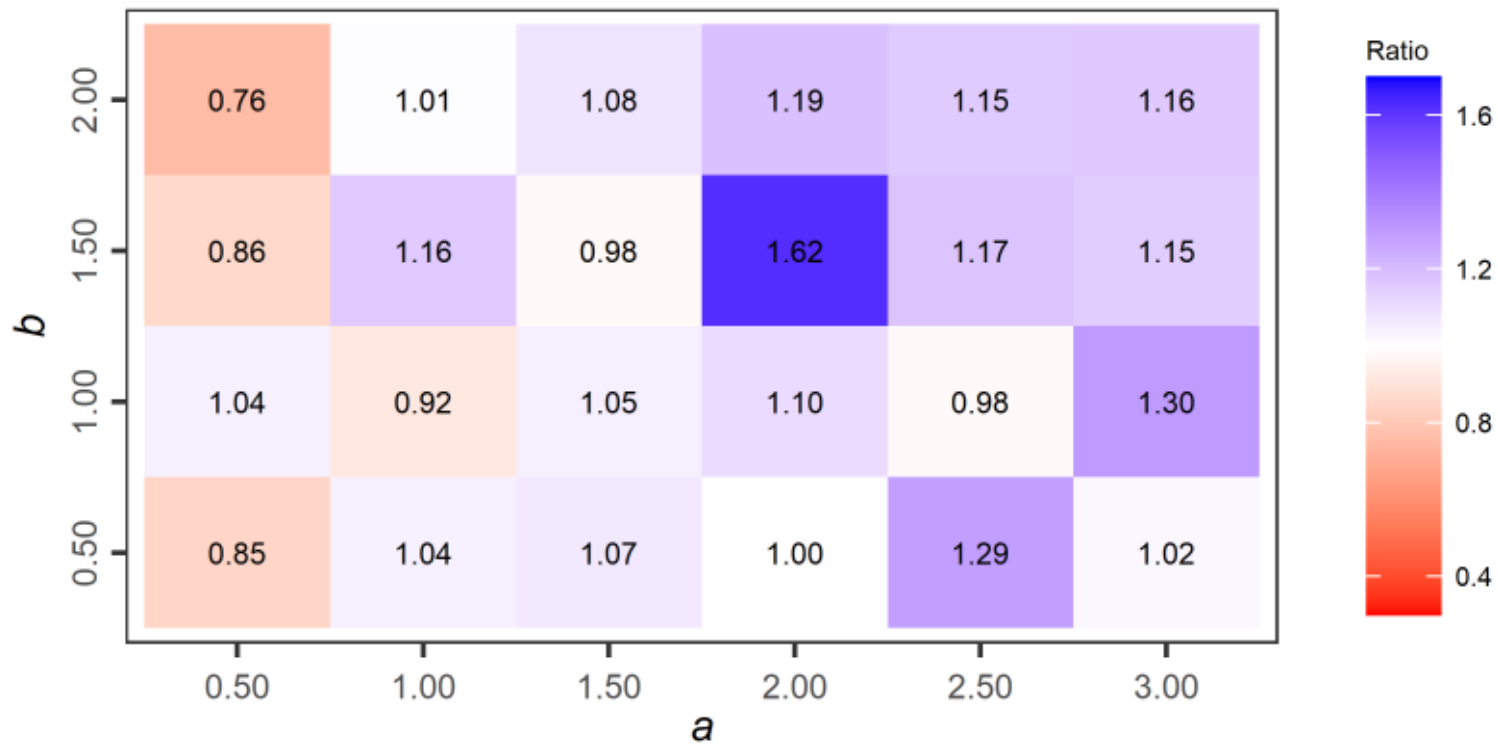

Figure 11. Average ratios $H_{a,b}^{\tau}(F)/Q^{\tau}(F)$ of predicted Huber quantiles and predicted quantiles at level $\tau = 0.5$ for all combinations of the parameters $a \in \{0.5, 1, 1.5, 2, 2.5, 3\}$ and $b \in \{0.5, 1, 1.5, 2\}$ of the Huber quantile.

### 4.3 Boston house prices (Supporting application)

A supporting application with the same three DL architectures (see Section 3.2) predicts Huber quantiles of house prices in Boston. This application intends to show the generality of DHQRN to other cases. Therefore, a detailed description of the dataset is out of scope for this study. The Boston house dataset is available for download from Fortmann-Roe (2015a, 2015b). The dataset includes 506 houses and we split the dataset randomly to three sets, i.e. a training set with 202 houses (40% of the samples), a validation set with 152 houses (30% of the samples) and a test set with 152 houses (30% of the samples). We selected 12 predictors, i.e. (1) per capita crime rate by town, (2) proportion of residential land zoned for lots over 25 000 sq.ft., (3) proportion of non-retail business acres per town, (4) Charles River dummy variable (= 1 if tract bounds river; 0 otherwise), (5) nitrogen oxides pollutant concentration (parts per 10 million), (6) average number of rooms per dwelling, (7) proportion of owner-occupied units built prior to 1940, (8) weighted distances to five Boston employment centres, (9) index of accessibility to radial highways, (10) full-value property-tax rate per ten thousand dollar, (11) minority: $1\,000(\mathrm{AM} - 0.63)^2$ where AM is the proportion of African Americans by town, and (12) percent lower status of the population. The dependent variable is median value of owner-occupied homes in thousands of dollars.

The parameters of the Huber quantiles were set to $p = 0.50$, $a = 5$, and $b = 4$. Note that the parameters $a$ and $b$ have the same units as the dependent variable (i.e., thousands of dollars). Model #1 had an average Huber quantile score of 3.99 and a Huber quantile level sample estimate of 0.50. Model #2 had an average Huber quantile score of 7.38 and a Huber quantile level sample estimate of 0.26. Model #3 had an average Huber quantile score of 5.79 and a Huber quantile level sample estimate of 0.40. Model #1 outperformed the remaining models with respect to the average score by a large margin. This is also reflected in the Huber quantile level sample estimates, which were perfect for Model #1 (i.e., equal to the nominal level of 0.50) but not satisfactory for the remaining two models.

We do not delve into the results in detail, as this application is intended to complement the primary application. The relative performance of the three algorithms differs from the previous application, consistent with the theoretical literature that states there is no algorithm universally superior to all others (Wolpert 1996). This holds true for both probabilistic predictions and predictions concerning functionals (Papacharalampous et al. 2024, Tyralis and Papacharalampous 2024).

### 4.4 Simulation example

To support the findings of the study, we also provide a simulation example as supplementary material. This includes a code in the `R` programming language (`Simulation_example_for_DHQRN.Rmd` file) as well as the results of the code (`Simulation_example_for_DHQRN.html` file). To define and train the neural network models, we used the `torch` (Falbel and Luraschi 2024) and `luz` (Falbel 2023) `R` packages.

We simulated 40 000 samples from the linear model $y = x + \underline{\varepsilon}$, where the error term $\underline{\varepsilon}$ follows a log-normal distribution with a mean and standard deviation of 0 and 1, respectively, on the log scale. The training set includes 30 000 samples, and the test set includes 10 000 samples. We trained a neural network with a single dense layer of 16 units, using:

a. The quantile scoring function (QRNN) with $\tau = 0.8$.

b. The expectile scoring function (ERNN) with $\tau = 0.8$.

c. The Huber quantile scoring function (DHQRN) with $\tau = 0.8, a = 3, b = 4$.

Remaining details of the trained neural networks are available in the code and report supplementary files and are omitted here for brevity. In three cases, the sample estimates of the levels at the test set were equal to $\tau = 0.7932$ (for QRNN), $\tau = 0.7999979$. (for ERNN), $\tau = 0.8099944$ (for DHQRN), Thus, they approximated the nominal level well.

We note again that in all cases, QRNN, ERNN, and DHQRN predict different functionals; therefore, they are not directly comparable, although QRNN and ERNN are edge cases of DHQRN. In our specific case, a linear model would outperform the neural network models, since the simulation data originated from a linear model. We believe that our code would be useful to the reader when aiming to customize it for more complex models.

## 4.5 Additional real-world applications

Additional real-world applications are presented below to demonstrate the satisfactory absolute performance of DHQRN. Table 2 presents the datasets, including details such as sample size, dependent and predictor variables, and data sources. To define and train the neural network models, we again utilized the `torch` (Falbel and Luraschi 2024) and `luz` (Falbel 2023) `R` packages. The codes for these real-world applications are available as supplementary material.

Table 2. Details of the datasets of additional real-world applications.

| Dataset | Sample size | Dependent variable | Predictor variables | Source (`R` package) |
|---|---|---|---|---|
| barro | 161 | y.net | lgdp2, fse2, gedy2 | `quantreg` |
| dutchboys | 6 848 | hgt | age, wgt, hc | `expectreg` |
| Gasoline | 342 | lgaspcar | lincomep, lrpmg, lcarpcap | `expectreg` |
| india | 4 000 | stunting | cbmi, cage, mbmi, mage | `expectreg` |

We compared the following two DL architectures:

- Model #1: A dense layers with `d_hidden1` units followed by a dropout layer with dropout rate equal to 0.5.

- Model #2: Two dense layers with `d_hidden1` and `d_hidden2` units respectively, each one followed by a dropout layer with dropout rate equal to 0.5.

Predictor variables were normalized using standard methods, while we were attentive to avoid information leakage from validation and test sets. Each dataset was randomly divided into three subsets: a training set (50% of the samples), a validation set (25% of the samples), and a test set (25% of the samples). Normalization was performed on both the training set and the combined training-validation set.

For training each model, we employed ReLU activation functions (Schmidt-Hieber 2020), the ADAM optimizer (Kingma and Ba 2015) with a learning rate `lr`, and set the number of training epochs to `n_epochs` and the batch size to 32 (Smith et al. 2018).

Specifically, the model was initially trained on the normalized training set. Optimal hyperparameters were then determined using the non-normalized validation set. Subsequently, the model parameters were re-estimated using the optimal hyperparameters on the normalized training-validation set. Finally, the performance of the model was evaluated on the non-normalized test set.

Hyperparameter tuning was conducted by creating a grid of all possible combinations of `d_hidden1`, `lr`, and `n_epochs` from Table 3 for Model #1. For Model #2, a grid of all possible combinations of `d_hidden1`, `d_hidden2`, `lr`, and `n_epochs` was created, with the constraint `d_hidden2` ≤ `d_hidden1` as specified in Table 3.

Table 3. Values of hyperparameters for creating grids for hyperparameter tuning.

| `d_hidden1` | `d_hidden2` | `lr` | `n_epochs` |
|---|---|---|---|
| 16 | 16 | 0.001 | 50 |
| 32 | 32 | 0.005 | 100 |
| | | 0.01 | |

Each model was trained using the generalized Huber scoring function, defined by eq. (24), with parameters $a$ and $b$ as specified in Table 4 for each dataset. It is important to note that parameters $a$ and $b$ must be pre-specified, because they represent the Huber functional that the modeler aims to predict. In all cases, we focused on Huber quantiles at the level $\tau = 0.75$.

Table 4. Definition of Huber functional parameters and sample levels of predictions on the test set by Models #1 and #2.

| Dataset | Huber functional parameters | Model #1 | Model #2 |
|---|---|---|---|
| barro | $a = 0.07, b = 0.06$ | 0.722 | 0.661 |
| dutchboys | $a = 100, b = 110$ | 0.717 | 0.718 |
| Gasoline | $a = 1, b = 1.2$ | 0.716 | 0.690 |
| india | $a = 200, b = 220$ | 0.750 | 0.751 |

Detailed results from all experiments are presented in reports in the supplementary material. Here, we report sample levels of the predictions on the test set for each model. Overall, the model performances are satisfactory. Model #1 appears to outperform Model #2 in most cases, based on absolute performance, as its sample levels are closer to the nominal level of 0.75; see Table 5.

## 5. Concluding remarks

We introduced deep Huber quantile regression networks (DHQRN) which is a wide class of deep learning (DL) regression networks that includes the cases of quantile regression neural networks (QRNN) and expectile regression neural networks (ERNN) as edge cases. The main concept for developing DHQRN is to train a DL algorithm using the Huber quantile scoring function as the objective function. DHQRN can predict Huber quantiles that are generalizations of quantiles and expectiles.

An example of house price prediction demonstrates the utility of the new algorithm. The predictions are then interpreted economically, based on mixture representations of the Huber quantile scoring function. The example also focuses on the relationship between DHQRN and its edge cases, by investigating their predictions.

Huber quantile regression is a non-parametric regression algorithm that has advantages over distributional regression, which requires the specification of a full conditional probability distribution (Waldmann 2018). If we are confident in the specification of the predictive probability distribution, the major advantage is that extrapolation is possible when using parametric methods. Otherwise, possible misspecification of the predictive distribution may result to considerable downgrading of predictive performance, thus quantile regression and its extensions should be preferred.

It is a common feature of machine learning studies to compare algorithms using multiple metrics. For instance, two algorithms are optimized using the Huber loss, but then they are compared using the squared error scoring function as well as the absolute error scoring function. The Huber loss is an approach for robust estimation of the location parameter (Huber 1964). In particular, asymptotic theory of estimating the location parameter for contaminated Gaussian distributions is developed by Huber (1964). As long as the predictive distribution is known to be Gaussian, the mean, median and Huber mean functionals are identical. Consequently, it makes sense to use the respective consistent scoring functions for estimation and predictive assessment reasons interchangeably.

However, in most cases, the predictive distribution is unknown, thus it may be assumed that it belongs to a wide family of distributions (e.g. distributions with finite first moments). In this case, it is not appropriate to use different scoring functions to assess predictions (Gneiting 2011) and the utility of the Huber quantile scoring function, should not be associated to its robustness, but to the specific characteristics of the respective elicitable predictive functionals. To assess, robustness properties of the Huber quantile scoring function, a way forward is to establish relationships between quantiles, expectiles and Huber quantiles for specific probability distributions and examine the effects of contamination.

Extension of the DHQRN to other types of DL models (beyond the densely-connected neural networks layers of the present study) is straightforward. Other types of data (e.g. energy and environmental data) could also be examined, with interpretation of results being a problem that requires critical attention.

Large-scale comparisons of quantile regression algorithms, such as quantile regression networks, quantile boosting regression, and quantile regression forests, have become common in the literature (e.g., Papacharalampous et al. 2024) and machine learning competitions. These comparisons aim to facilitate understanding of the finite sample properties of these algorithms in various situations. Therefore, extensions of Huber quantile regression to other types of ML algorithms (boosting algorithms (Friedman 2001, Tyralis and Papacharalampous 2021a) quantile regression forests (Meinshausen 2006)) combined with large scale comparisons between different Huber quantile regression algorithms is a way to move forward. The case of predictions of extremes beyond using semi-parametric methods (e.g. see the application Tyralis and Papacharalampous 2023) is also of interest (see e.g. advancements in Taggart 2022a).

## Appendix A Statistical software

We used the `R` programming language (`R` Core Team 2023) and `Python` programming language (Van Rossum and Drake 2009) to implement the algorithms, and to report and visualize the results.

For data processing and visualizations, we used the contributed `R` packages `A3` (Fortmann-Roe 2015a, 2015b), `data.table` (Dowle and Srinivasan 2023), `ggmap` (Kahle and Wickham 2013, Kahle et al. 2023), `MASS` (Ripley 2023, Venables and Ripley 2002), `reticulate` (Kalinowski et al. 2023), `tidyverse` (Wickham et al. 2019, Wickham 2023) and the contributed `Python` libraries `numpy` v.1.23.5 (Harris et al. 2020), `pandas` v.1.5.3 (McKinney 2010, The pandas development team 2023), `scikit-learn` v.1.2.2 (Pedregosa et al. 2011), `seaborn` v.0.12.2 (Waskom 2021).

DL algorithms of the real case studies were developed with the contributed `Python` library `tensorflow` v.2.12.0 (Abadi et al. 2016a, 2016b, TensorFlow Developers 2023).

Dl algorithhms of the simulation example were developed with the contributed `R` packages `luz` (Falbel 2023), `torch` (Falbel and Luraschi 2024).

The performance metrics were computed by implementing the contributed `R` package `scoringfunctions` (Tyralis and Papacharalampous 2022, 2024).

Data for the real-world applications of Section 4.5 were retrieved from the contributed `R` package `expectreg` (Otto-Sobotka et al. 2024, Schnabel and Eilers 2009, Sobotka and Kneib 2012), `quantreg` (Koenker 2024).

Reports were produced by using the contributed `R` packages `devtools` (Wickham et al. 2022), `knitr` (Xie 2014, 2015, 2023), `rmarkdown` (Allaire et al. 2023, Xie et al. 2018, 2020).

**Supplementary material:** We provide the `r_functions.R`, `Simulation_example_for_DHQRN.Rmd`, `real_example_1_barro_dataset.Rmd`, `real_example_2_dutchboys_dataset.Rmd`, `real_example_3_Gasoline_dataset.Rmd`, `real_example_4_india_dataset.Rmd` code files, as well as the `Simulation_example_for_DHQRN.html`, `real_example_1_barro_dataset.html`, `real_example_2_dutchboys_dataset.html`, `real_example_3_Gasoline_dataset.html`, `real_example_4_india_dataset.html` report files.

**Acknowledgement:** We thank the Associate Editor and reviewers whose comments contributed to improving the manuscript.

**Conflicts of interest:** The authors declare no conflict of interest.

**Author contributions:** HT and GP conceptualized and designed the work with input from ND and KPC. HT and GP performed the analyses and visualizations, and wrote the first draft, which was commented and enriched with new text, interpretations and discussions by ND and KPC. ND and HT designed the graphical abstract.